\documentclass[10pt,twocolumn,letterpaper]{article}

\usepackage[pagenumbers]{wacv} 

\usepackage{graphicx}
\usepackage{amsmath}
\usepackage{amssymb}
\usepackage{booktabs}
\usepackage{xcolor}
\usepackage{multirow}
\usepackage{colortbl}
\usepackage{hhline}
\definecolor{Cerulean}{rgb}{0,0,0.95}
\definecolor{LimeGreen}{rgb}{0.15,0.65,0.15}
\definecolor{RoyalBlue}{rgb}{0.25,0.41,0.88}
\definecolor{Rose}{rgb}{1.0, 0.15, 0.21}
\definecolor{Orange}{rgb}{1.0, 0.5, 0.0}
\definecolor{Gray}{gray}{0.6}
\definecolor{Black}{gray}{0.0}
\definecolor{Purple}{rgb}{0.77,0.12,0.64}

\newcommand{\authors}[1]{{\color{Black} #1}}

\usepackage{url}
\usepackage[pagebackref,breaklinks,colorlinks]{hyperref}
\usepackage[hyphenbreaks]{breakurl}
\usepackage[accsupp]{axessibility} 

\usepackage[capitalize]{cleveref}
\crefname{section}{Sec.}{Secs.}
\Crefname{section}{Section}{Sections}
\Crefname{table}{Table}{Tables}
\crefname{table}{Tab.}{Tabs.}

\def\wacvPaperID{11} 
\def\confName{WACV}
\def\confYear{2024}

\begin{document}

\title{Human-Centric Autonomous Systems With LLMs for User Command Reasoning}

\author{Yi Yang$^{1,2}$, Qingwen Zhang$^1$, 
Ci Li$^1$, Daniel Simões Marta$^1$, Nazre Batool$^2$, John Folkesson$^1$ \\
$^1$ KTH Royal Institute of Technology \quad  $^2$ Scania AB \\
{\tt\footnotesize \{yiya, qingwen, cil, dlmarta, johnf\}@kth.se} ~~ {\tt\footnotesize \{carol-yi.yang, nazre.batool\}@scania.com}
}

\maketitle

\begin{abstract}
\authors{The evolution of autonomous driving has made remarkable advancements in recent years, evolving into a tangible reality. However, a human-centric large-scale adoption hinges on meeting a variety of multifaceted requirements.}
\authors{To ensure that the autonomous system meets the user's intent, it is essential to accurately discern and interpret user commands, especially in complex or emergency situations.}
\authors{To this end, we propose to leverage the reasoning capabilities of Large Language Models (LLMs) to infer system requirements from in-cabin users' commands. Through a series of experiments that include different LLM models and prompt designs, we explore the few-shot multivariate binary classification accuracy of system requirements from natural language textual commands. We confirm the general ability of LLMs to understand and reason about prompts but underline that their effectiveness is conditioned on the quality of both the LLM model and the design of appropriate sequential prompts.}
\authors{Code and models are public with the link \url{https://github.com/KTH-RPL/DriveCmd_LLM}.}

\end{abstract}

\section{Introduction}
\label{sec:intro}
\authors{Autonomous driving (AD) has experienced significant advances in recent years~\cite{Geiger2013IJRR, sun2020scalability, liu2021role}, enhanced by advancements in both hardware and machine learning techniques. 
The field has seen diverse developments: multi-task learning approaches~\cite{chen2022lav} over traditional modular systems~\cite{gog2021pylot}, end-to-end methodologies~\cite{zhang2022mmfn, jia2023think} with planning-oriented techniques~\cite{hu2023planning, jia2023driveadapter}, and widespread adoption of Bird-Eye-View representations~\cite{li2022bevformer}.
Many explorations have been conducted towards safe and robust autonomous systems. 

Nonetheless, several aspects are often overlooked, such as taking into consideration human intent when designing AD systems~\cite{chen2022milestonessurvey}.}  The future of AD should integrate human-centered design with advanced AI capabilities to reason and interpret the user's intent~\cite{cui2023drive,cui2023receive}. This integration offers numerous advantages e.g. language interaction~\cite{wayve_LINGO}, driving scene understanding~\cite{drivelm2023}, contextual reasoning~\cite{yao2022react}, as well as explainability and trust.
In a recent milestone survey, Chen et al.~\cite{chen2022milestonessurvey} emphasize the necessity of considering human behaviors and AD systems to ensure communication transparency and efficiency. They also hint towards a human-machine hybrid intelligence approach, asserting that the reliability of intelligent systems~\cite{zheng2017hybrid} depends on incorporating learnability and the influence of human intelligence and mentorship.

A current relevant challenge stems from allowing humans to interact using natural language with AD systems~\cite{drivelm2023, fu2023drive}, aiming to adhere to both human preferences and intent. Our contribution here is a series of insights and experiments on a variety of LLM models and prompt designs exploring the interaction between users and AD systems via verbal commands.

While the use of natural language inherently poses a challenge in itself, we leverage recent advancements in large pre-trained foundational models~\cite{bommasani2021opportunities, brown2020language, openai2023gpt4} alongside their zero-shot capabilities to adapt to new tasks~\cite{zeng2022socratic}.
In our approach, we adopt a few-shot learning strategy, keeping an LLM frozen and enhancing reasoning through sequential prompting, drawing inspiration from previous works~\cite{wei2022chain,cui2023drive,cui2023receive}.
Ultimately, our goal is to condition an LLM for the task of multivariate binary classification of AD systems based on in-cabin user commands.

\begin{table*}
  \begin{minipage}{0.99\textwidth}
\centering  
\vspace{-4mm}
\scalebox{0.88}{
\begin{tabular}{l p{15.5cm} }
\toprule
 \multicolumn{2}{l}{\bf LLM prompt conditioning and prompt design with user examples}  \\
\midrule

a) LLM Conditioning & \scriptsize{\texttt{{You'll receive a command message for a self-driving vehicle. Follow these steps to respond:}}}\\
 & \scriptsize{\texttt{\color{Orange} Step 1: \color{Gray}First decide whether the external perception system is required for this command. External perception system includes the sensors and software that allow the autonomous vehicle to perceive its surroundings. 
It typically includes cameras, lidar, radar, and other sensors to detect objects, pedestrians, other vehicles, road conditions, and traffic signs/signals. For example, any movement, sense or detect the surroundings.}} \\ 
&  \scriptsize{\texttt{\color{Orange}Step 2: \color{Gray} Answer "Is in-cabin monitoring required?"
in-cabin monitoring involves cameras, thermometers, or other sensors placed inside the vehicle’s cabin to monitor the state of occupants and other conditions. 
It includes everything in-cable system, for example, seats, windows, doors, multimedia system, alert system, etc.}} \\
& ... ...\\
& \scriptsize{\texttt{\color{Orange}Step 8: \color{Gray} Answer "Is there a possibility of violating traffic laws?"
Violating traffic laws refers to any action performed by the vehicle that goes against the established traffic regulations of the region. An autonomous vehicle’s system is typically designed to adhere strictly to traffic laws. It includes anything risk command. For example, related to the traffic laws, like speed, traffic light, emergency action etc.}}
\\
& \scriptsize{\texttt{Answer the 8 questions use the following format: }} \\
 & \scriptsize{\texttt{\color{Orange}Step 1: `Yes' or `No' <step 1 reasoning>}} \\
 & \scriptsize{\texttt{\color{Orange}Step 2: `Yes' or `No' <step 2 reasoning>}} \\
 & ... ... \\
& \scriptsize{\texttt{\color{Orange}Step 8: `Yes' or `No' <step 8 reasoning> }} \\
& \scriptsize{\texttt{\color{LimeGreen}Response to user: Output is [A1 A2 A3 A4 A5 A6 A7 A8]. 
Replace A1-A8 with 1 for `Yes' and 0 for `No'.}} \\
\midrule
 & [{\it Provide few-shots for LLM to learn. Show one shot as an example below.}] \\
b) User & \scriptsize{\texttt{\color{Purple}Call my friend Carol.}} \\
Assistant & \scriptsize{\texttt{\color{RoyalBlue}Explaination:~\color{Gray}}} \\
& \scriptsize{\texttt{\color{Orange} Step 1: \color{Gray}
No, it doesn't involve any movement, sense or detect the surrounding.}} \\
& \scriptsize{\texttt{\color{Orange}Step 2: \color{Gray}
Yes, it requires to use the in-cabin multimedia to call the people.}} \\
 & ... ... \\
& \scriptsize{\texttt{\color{Orange}Step 8: \color{Gray} No, it is not a risky command.}} \\
& \scriptsize{\texttt{\color{LimeGreen}Therefore, the output should be : [0 1 0 0 1 1 1 0].}}
\\
\midrule
 & [{\it Test on real commands.}] \\
c) User & \scriptsize{\texttt{\color{Purple}Make an illegal U-turn here.}} \\
GPT-4~\cite{openai2023gpt4} &  
\scriptsize{\texttt{\color{RoyalBlue}Explanation:~\color{Gray}}} \\
& \scriptsize{\texttt{\color{Orange} Step 1: \color{Gray}
Yes, it needs to sense or detect the surrounding for making a u-turn.}} \\
& \scriptsize{\texttt{\color{Orange}Step 2: \color{Gray}
No, it doesn't involve anything inside the vehicle's in-cabin system, like seats, windows, doors, multimedia system, alert system, etc. }} \\
& ... ... \\
& \scriptsize{\texttt{\color{Orange}Step 8: \color{Gray} Yes, it is a risky command and it is illegal to make a u-turn}} \\
& \scriptsize{\texttt{\color{LimeGreen}Therefore, the output should be : [1 0 1 1 0 0 0 1].}}
\\
\bottomrule
\end{tabular}
}
\vspace{2mm}
\caption{Macroscopic view of our approach. For visual readability, steps are in \textcolor{Orange}{orange} (each step contains one question), user inputs are in \textcolor{Purple}{purple}, explanations are in \textcolor{RoyalBlue}{blue}, and the corresponding LLM classification output is in \textcolor{LimeGreen}{green}. a) LLM initial conditioning of several steps in the form of a prompt.
b) Few-shot examples resembling a \emph{chain-of-thought}~\cite{wei2022chain} to improve reasoning. c) Example with test commands on GPT-4.}  
\label{tab:user_example}  
  \end{minipage}
\end{table*}

\section{Related Work}
\authors{\textbf{Few-shot Reasoning Capabilities of LLMs.} Recent research indicates that LLMs with sufficient expansiveness possess the ability to perform sophisticated reasoning tasks~\cite{ho2022large,zeng2022socratic}. In this work, we adopt an approach somewhat akin to Socratic models~\cite{zeng2022socratic}, where models--encompassing vision, language, and sound--can operate in a zero-shot or few-shot fashion. A prevalent strategy in this type of approach involves keeping certain model subcomponents static, particularly those relevant to a single modality, preserving them for use in subsequent tasks~\cite{tsimpoukelli2021multimodal,NEURIPS2022_215aeb07}. This methodology interchangeably aligns well with few-shot transfer learning~\cite{gavves2015active,schwab2018zero,ying2018transfer,soh2020meta}, wherein an LLM, originally trained on a vast array of internet-scale text prompts for various tasks such as text completion and sentiment analysis, is repurposed for a new, specific task. In our case, this involves leveraging general domain knowledge of AD for a classification challenge~\cite{nag2021towards}. However, the success of adapting language models to new tasks largely depends on the prompting strategy and the quality of the prompts~\cite{lester2021power}. Wei et al.~\cite{wei2022chain} demonstrated that standard prompting is often insufficient. They introduced a \emph{chain-of-thought} prompting strategy, where a sequence of thoughtful demonstrations significantly improves performance in tasks such as symbolic reasoning, commonsense, and arithmetic. Inspired by the above works, we aim to keep the weights of an LLM frozen and design a chain-of-thought strategy for a downstream task of classification from in-cabin user commands.}

\begin{table*}[t!]
\centering
\def\arraystretch{1.2}
\resizebox{\textwidth}{!}{%
\begin{tabular}{lcccccccccc} 
\toprule
\multicolumn{1}{c|}{\multirow{4}{*}{Methods}} & \multicolumn{10}{c}{Accuracy (\%) ↑}                                                                                                                                                                                                                                                                                                                                                                                                                                                                                                    \\ 
\cline{2-11}
\multicolumn{1}{c|}{}                         & \multicolumn{1}{c|}{\multirow{3}{*}{\begin{tabular}[c]{@{}c@{}}Command\\Level\end{tabular}}} & \multicolumn{9}{c}{Question Level}                                                                                                                                                                                                                                                                                                                                                                                                     \\ 
\cline{3-11}
\multicolumn{1}{c|}{}                         & \multicolumn{1}{c|}{}                                                                        & Overall                                   & Perception & \begin{tabular}[c]{@{}c@{}}In-cabin\\Monitoring\end{tabular} & Localization & \begin{tabular}[c]{@{}c@{}}Vehicle\\control\end{tabular} & \begin{tabular}[c]{@{}c@{}}Entertain-\\ment\end{tabular} & \begin{tabular}[c]{@{}c@{}}Personal\\Data\end{tabular} & \begin{tabular}[c]{@{}c@{}}Network\\Access\end{tabular} & \begin{tabular}[c]{@{}c@{}}Traffic\\laws\end{tabular}  \\ 
\midrule
Random                                        & {\cellcolor[rgb]{0.949,0.949,0.949}}0.36                                                     & {\cellcolor[rgb]{0.949,0.949,0.949}}49.44 & 50.59      & 47.68                                                        & 51.96        & 48.95                                                    & 47.86                                                    & 48.23                                                  & 50.59                                                   & 49.68                                                  \\
Rule-based*                                   & {\cellcolor[rgb]{0.949,0.949,0.949}}4.09                                                     & {\cellcolor[rgb]{0.949,0.949,0.949}}69.19 & 62.33      & 74.52                                                        & 69.70        & 65.15                                                    & 87.99                                                    & 70.43                                                  & 58.60                                                   & 64.79                                                  \\ 
\hline
CodeLlama-34b-Instruct                        & {\cellcolor[rgb]{0.949,0.949,0.949}}16.28                                                    & {\cellcolor[rgb]{0.949,0.949,0.949}}74.56 & 65.61      & 68.61                                                        & 78.53        & 79.71                                                    & 75.98                                                    & 70.15                                                  & 75.07                                                   & 82.80                                                  \\
Llama-2-70b-Chat                              & {\cellcolor[rgb]{0.949,0.949,0.949}}21.29                                                    & {\cellcolor[rgb]{0.949,0.949,0.949}}82.46 & 87.17      & 83.44                                                        & 90.35        & 87.81                                                    & 79.98                                                    & 77.07                                                  & 85.90                                                   & 67.97                                                  \\ 
\hline
GPT-3.5                                       & {\cellcolor[rgb]{0.949,0.949,0.949}}\underline{36.21}                                                    & {\cellcolor[rgb]{0.949,0.949,0.949}}\underline{88.03} & 88.63      & 86.44                                                        & 90.17        & 87.44                                                    & 96.09                                                    & 80.98                                                  & 87.90                                                   & 86.62                                                  \\
GPT-4                                         & {\cellcolor[rgb]{0.949,0.949,0.949}}\textbf{38.03}                                                    & {\cellcolor[rgb]{0.949,0.949,0.949}}\textbf{89.02} & 93.18      & 74.89                                                        & 91.54        & 88.63                                                    & 94.45                                                    & 85.99                                                  & 91.99                                                   & 91.54                                                  \\
\bottomrule
\end{tabular}
}
\caption{Performance comparison of different methods or models on the user command benchmark regarding 8 yes/no questions. We compare with the methods of random guess and rule-based as baseline. *Note that the rules are generated by the GPT-4 with Advanced Data Analysis plug-in. With our designed prompt shown in \cref{tab:user_example}, we compare several LLMs including Llama, CodeLlama, GPT-3.5-turbo, and GPT-4. The best results are shown in \textbf{bold} and the second best results are shown in \underline{underlined}.}
\label{tab:main_res}
\end{table*}
\textbf{Human-centric Autonomous Driving.} Incorporating human intent in the form of natural language within AD is a relatively new and active field of research. The emergence of recent datasets and works has significantly boosted its development.
Datasets like NuPrompt~\cite{wu2023language}, NuScenes-QA~\cite{qian2023nuscenes}, DRAMA~\cite{malla2023drama}, enhance the autonomous driving datasets with provided texts, for different tasks including object tracking, visual question answering, image caption and so on.
As many works have shown promising results,
a possible goal would be to mediate planning-oriented AD~\cite{hu2023planning}, which includes perception, prediction and planning, with human intent. In recent exploratory work, there is verified evidence for the effectiveness of integrating LLMs into human-centric AD ~\cite{jin2023adapt, mrazovac2022human,li2021human, Cui_2024_WACV}. 
In a survey by Zhou et al.~\cite{zhou2023vision}, large language and vision models show potential to contribute to the AD system in different submodules including perception and understanding, navigation and planning, decision-making and control, as well as end-to-end pipelines.
For instance, as explored by Fu et al.~\cite{fu2023drive}, understanding the common sense driving intentions embedded in human commands is achieved by utilizing the reasoning capacities of LLMs in addressing long-tailed cases. This type of approach could be instrumental in the evolution of AD technologies that mirror human-like driving nuances. 
Ding et al.~\cite{ding2023hilm} use multimodal LLM to inform AD system the localization of risk objects and provide suggestions for safety and robustness.
More closely related to our work is that reported by Cui et al.~\cite{cui2023drive,cui2023receive} and  Jain et al.~\cite{jain2023ground}, demonstrating that utilizing linguistics and understanding from verbal commands can improve driving decisions and enhance personalized driving experiences through ongoing verbal feedback. In this exploratory work, we aim to holistically address some of the weaknesses posed when handling long-tail data or out-of-distribution driving scenarios by tapping into the overall implicitly acquired domain knowledge of LLMs~\cite{zhao2021explainable,nag2021towards}.

\section{Method}
\label{sec:methods}
Given user commands as input text sentences, the model is expected, through reasoning and sentiment analysis, to provide binary results (`Yes' / `No') for 8 specific questions. These questions inquire about the command's diverse requirements- ranging from perception, in-cabin monitoring, localization, control, network access, and entertainment to human privacy and traffic laws.

Since LLMs undergo training on extensive datasets, integrating them within autonomous systems significantly enhances AD systems' capability to grasp both scene dynamics and user intent. In this paper, we utilize both online (from the GPT series \cite{openai2023gpt4}) and offline (from the Llama series \cite{touvron2023llama}) LLMs. As demonstrated in \cref{tab:user_example}, we begin by informing the LLM that its task is to engage in a question and answer (Q\&A) format by responding with `Yes' or `No'. By instructing the model with an in-depth explanation for each of the eight questions, it is prompted to process information step by step. To further optimize accuracy, the LLMs are provided with few examples of in-context few-shot learning \cite{brown2020language}. Finally, conditioned on the real user command, LLMs are instructed to produce both step-by-step explanations and results in a predetermined format. 

\section{Experiment}
\label{sec:expertiments}
\textbf{Datasets and Metrics}:
We assess the performance of LLMs on 1,099 in-cabin user commands from the UCU Dataset~\cite{ucu_dataset}. The details are available on the challenge website\footnote{\url{https://llvm-ad.github.io/challenges/}}. The evaluation involves determining if a command requires any of the eight specified modules' help to achieve autonomy. Official evaluation metrics include accuracy at \textit{the question level} (accuracy for each individual question) and at \textit{the command level} (accuracy is only acknowledged if all questions for a particular command are correctly identified).

\textbf{LLM Models and Baseline}:
We benchmark our prompt approaches using a range of LLMs, including GPT-3.5-turbo / GPT-4\footnote{ \url{https://platform.openai.com}. Note that during the time we used the models, GPT-3.5-turbo pointed to GPT-3.5-turbo-0613 version, and GPT-4 pointed to GPT-4-0613 version.},
and CodeLlama-34b-Instruct / Llama-2-70b-Chat~\cite{touvron2023llama}. The GPT models are accessed online, whereas Llama models offer an on-board solution, with both their code and pretrained weights open-sourced. CodeLlama-34b-Instruct is an enhanced version of the original Llama \cite{llama1} additionally trained on code generation and instruction problems with 34 billion parameters in the model. Llama-2-70b-Chat has roughly double the number of parameters and is trained mainly on conversational interactions enhanced with human reinforcement learning.

We compare the LLMs' performance against two baselines. The first employs a simple random guessing strategy. The second is a rule-based method that identifies specific keywords in the user command to determine system requirements. It is noteworthy that these rules are automatically generated by ChatGPT-4, augmented with the Advanced Data Analysis plug-in.
\begin{table}
\centering
\def\arraystretch{1.2}
\resizebox{\linewidth}{!}{%
\begin{tabular}{cc|c|cc} 
\toprule
\multicolumn{2}{c|}{Detailed Explanation}                    & \multicolumn{1}{c|}{\multirow{2}{*}{Few Shots}} & \multicolumn{2}{c}{Level-based Accuracy (\%) ↑}                                       \\ 
\cline{1-2}\cline{4-5}
\multicolumn{1}{l|}{w/o step} & \multicolumn{1}{l|}{w step} & \multicolumn{1}{c|}{}                           & \multicolumn{1}{c}{Command} & \multicolumn{1}{c}{Question}  \\ 
\midrule
                          &            &                                                 & 13.00                            & 75.00                               \\
\checkmark &                           &                                                 & 22.00                             & 81.00                               \\
                          & \checkmark &                                                 & 26.00                          & 79.00                               \\
                          &                           & \checkmark                       & 35.00                            & 87.00                               \\
\checkmark &                           & \checkmark                       & \textbf{42.00}                            & \textbf{88.63}                         \\
                          & \checkmark & \checkmark                       & \underline{40.00}                           &\underline{87.88}                              \\
\bottomrule
\end{tabular}

}
\caption{Ablation study in the impact of different prompts.}
\label{tab:ab_param}
\end{table}

\begin{table}
\centering
\def\arraystretch{1.2}
\resizebox{0.65\linewidth}{!}{%
\centering
\begin{tabular}{ccc} 
\toprule
\multirow{2}{*}{\# of shot} & \multicolumn{2}{c}{Level-based Accuracy (\%) ↑}                                       \\ 
\cline{2-3}
                            & \multicolumn{1}{c}{Command} & \multicolumn{1}{c}{Question}  \\ 
\hline
0                           & 26.00                            & 79.00                              \\
1                           & 23.00                            & 80.63                              \\
2                           & \underline{40.00}                             & \underline{87.63}                               \\
3                           & \textbf{41.00}                            & \textbf{87.88}                               \\
4                           & \underline{40.00}                             & \textbf{87.88}                              \\
\bottomrule
\end{tabular}
}
\caption{Ablation study in the number of provided shots.}
\label{tab:ab_num_shot}
\end{table}

\textbf{Evaluation}:\label{evaluation}
The results are presented in \cref{tab:main_res}. Compared to the random guess and rule-based approaches, LLMs exhibit notably higher accuracy, especially at the command level. 
Among the LLMs tested, the GPT series surpasses the Llama models, where GPT-4 achieves a peak accuracy of 89.02\% at the question level and 38.03\% at the command level. 
Diving into question-specific accuracy, GPT consistently behaves excellently in determining if a command requires perception, localization, entertainment submodules, or if it might violate traffic laws. However, for the in-cabin monitoring, GPT's predictions display some inconsistency and lower accuracy, especially for GPT-4 (74.89\%).
It is observed that in the evaluation of 1,099 commands, GPT-4 incorrectly responded to 276 queries related to in-cabin monitoring.
A majority proportion of these errors, amounting to 259 instances, involved GPT-4 saying ``Yes, it requires in-cabin monitoring" when the ground truth doesn't, with only a few errors being the reverse.
To further investigate why GPT-4 says yes often, 2 keywords are found in its explanations: \textit{multimedia} and \textit{alert} system.
Both terms are typically associated with in-cabin monitoring in our given instructed prompts.
Specifically, 155 commands (56\%) are related to multimedia - GPT-4 identifies these as requiring in-cabin multimedia activities, such as making calls, playing radio or videos, or screen displays. 
Additionally, in 26\% of the instances (71 commands), GPT-4 associates the command with the utilization of in-cabin monitoring systems for alerting or notifying the vehicle's occupants.
Given this context, it's difficult to conclusively label GPT-4's responses as incorrect. This is largely due to the vague definition of what exactly constitutes in-cabin monitoring in the context of our study.

\textbf{Ablation Study}:
We perform two ablation studies to explore the impacts of providing a detailed explanation of instructions and the given number of shot examples. 
To reduce costs, we run experiments on GPT-3.5 with a small test subset that matches the distribution of the original test data created by ChatGPT-4.
More specifically, ChatGPT-4 is asked to analyze the data distribution for each question and sample 100 commands according to it.

The performance of combining three prompting methods is shown in \cref{tab:ab_param}. 
Note that for providing a detailed instruction explanation, we experimented with two formats: a step-by-step approach as illustrated in \cref{tab:user_example}, and a consolidated paragraph format (labeled as `w/o step' in \cref{tab:ab_param}).
For the latter case, we omit the word \texttt{Step:\#} which causes less structured prompts.
Each prompting method exhibits enhanced performance, with the few-shot method outperforming two detailed explanation methods. Notably, combining one of the prompting methods from detailed explanation and the few-shot method significantly augments the performance. 

\cref{tab:ab_num_shot} shows the impact of the number of given few shot examples with the step-by-step detailed explanation method. We observe a notable increase when two examples are provided, which indicates the importance of providing shot examples for performance gain. However, as the number of examples continues to increase, the impact on the final results becomes subtle, suggesting the existence of a saturation threshold or limitations.

\section{Conclusion}
\label{sec:conclusion}
In this paper, we offered a human-centric perspective by providing several key insights with different prompting designs to condition LLMs towards achieving AD system requirements from verbal user commands. Our work included an analysis of several online and offline popular LLMs with a variety of ablation studies.
We hope that this work inspires our community to consider human intent in the design of an AD system.
Future work points towards the direction of incorporating human feedback in a more intelligent and effective way.  This would pave the way for the development of AD systems that are not only more reliable and capable in their reasoning and comprehension skills but also more closely aligned with human-like preferences and behaviors.
The goal is to establish an AD system with human-centric values, thereby fostering greater trust and acceptance among users.
\section*{Acknowledgement}
Thanks to RPL's members: Xuejiao Zhao and Boyue Jiang, who gave constructive comments on this work. 
We also thank the anonymous reviewers for their constructive comments.

This work\footnote{We have used ChatGPT for editing and polishing author-written text.} was funded by Vinnova and Wallenberg AI, Autonomous Systems and Software Program (WASP), Sweden (research grant). The computations were enabled by the supercomputing resource Berzelius provided by National Supercomputer Centre at Linköping University and the Knut and Alice Wallenberg foundation, Sweden.
{\small
\bibliographystyle{unsrt}
\bibliography{egbib}
}

\clearpage
\onecolumn
\appendix
\section{Qualitative Results of GPT4}
There are three examples of the GPT4 real results with explanations for different user commands in \cref{tab:good_sample1}, \cref{tab:good_sample2}, \cref{tab:bad_sample}.
\vspace{6em}
\begin{table*}[h]
  \begin{minipage}{0.99\textwidth}
\centering  
\vspace{-4mm}
\scalebox{0.88}{
\begin{tabular}{l p{15.5cm} }
\toprule
 \multicolumn{2}{l}{\bf Some user command examples}  \\
\midrule

a) LLM Conditioning & {\texttt{{[Set as in \cref{tab:user_example}.]}}}
\\
b) User Assistant & {\texttt{[Provide few-shots for LLM to learn shown in \cref{tab:user_example}.]}} \\
\midrule
 & [{\it Test on real commands.}] \\
c) User & {\texttt{\color{Purple}Share the vehicle's location to my dad.}} \\
GPT-4~\cite{openai2023gpt4} &  
{\texttt{\color{RoyalBlue}Explanation:~\color{Gray}}} \\
& \texttt{\textcolor{Orange}{
1. External Perception system:} Is it required?
}\\ & \texttt{
No, it doesn't involve any movement, sense or detect the surrounding.
}\vspace{0.5em}\\
& \texttt{\textcolor{Orange}{
2. In-cabin monitoring:} Is it required?
}\\ & \texttt{
No, it doesn't involve anything inside the vehicle cabin to monitor something.
}\vspace{0.5em}\\
& \texttt{\textcolor{Orange}{
3. Localization:} Is it required?
}\\ & \texttt{
Yes, it needs to know the self-driving vehicle's position.
}\vspace{0.5em}\\
& \texttt{\textcolor{Orange}{
4. Vehicle control:} Is it required?
} \\ & \texttt{
No, it doesn't involve physically controls for the vehicle movements, such as steering, acceleration, braking, and signaling.
}\vspace{0.5em}\\
& \texttt{\textcolor{Orange}{
5. Entertainment system:} Is it required?
} \\ & \texttt{
No, it is not about the entertainment at all.
}\vspace{0.5em}\\
& \texttt{\textcolor{Orange}{
6. User personal data:} Is it required?
} \\ & \texttt{
Yes, it involves one person's phone, which is an identifiable individual.
}\vspace{0.5em}\\
& \texttt{\textcolor{Orange}{
7. External network access:} Is it required?
}\\ & \texttt{
Yes, it belongs to the cases that need to search for information from the internet like some places, route path from one place to another,  contacts, weather, movie, music, etc. To get to know the nearest parking lot, it requires internet to search for it.
}\vspace{0.5em}\\
& \texttt{\textcolor{Orange}{
8. Traffic laws:} Is there a possibility of violating them?
}\\ & \texttt{
No, it should not involve in this case. As it is a normal driving command, which is not risky.
}\vspace{0.5em}\\
& \texttt{\textcolor{LimeGreen}{Therefore, the output should be} //[\colorbox{yellow}{0 0 1 0 0 1 1 0}]//}
\\
\bottomrule
\end{tabular}
}
\vspace{2mm}
\caption{High accuracy example 1. \colorbox{yellow}{Yellow} highlights the right answer. \colorbox{pink}{Red} highlights the wrong answer. In this command \textit{``Share the vehicle's location to my dad"}, all answers are correct.}  
\label{tab:good_sample1}  
  \end{minipage}
\end{table*}

\begin{table*}
  \begin{minipage}{0.99\textwidth}
\centering  
\vspace{-4mm}
\scalebox{0.88}{
\begin{tabular}{l p{15.5cm} }
\toprule
 \multicolumn{2}{l}{\bf Some user command examples}  \\
\midrule

a) LLM Conditioning & {\texttt{{[Set as in \cref{tab:user_example}.]}}}
\\
b) User Assistant & {\texttt{[Provide few-shots for LLM to learn shown in \cref{tab:user_example}.]}} \\
\midrule
 & [{\it Test on real commands.}] \\
c) User & {\texttt{\color{Purple}Take the kids to school and return home.}} \\
GPT-4~\cite{openai2023gpt4} &  
{\texttt{\color{RoyalBlue}Explanation:~\color{Gray}}} \\
& \texttt{\textcolor{Orange}{
1. External Perception system:} Is it required?
}\\ & \texttt{
Yes, it needs driving, to sense or detect the surroundings.
}\vspace{0.5em}\\
& \texttt{\textcolor{Orange}{
2. In-cabin monitoring:} Is it required?
}\\ & \texttt{
Yes, it involves the in-cabin monitoring to ensure the safety of the kids.
}\vspace{0.5em}\\
& \texttt{\textcolor{Orange}{
3. Localization:} Is it required?
}\\ & \texttt{
Yes, it needs navigation, planning route, and it is about from ego location to other destinations.
}\vspace{0.5em}\\
& \texttt{\textcolor{Orange}{
4. Vehicle control:} Is it required?
} \\ & \texttt{
Yes, it needs to control the vehicle's movement.
}\vspace{0.5em}\\
& \texttt{\textcolor{Orange}{
5. Entertainment system:} Is it required?
} \\ & \texttt{
No, it doesn't need multimedia system including radio, music, video, etc.
}\vspace{0.5em}\\
& \texttt{\textcolor{Orange}{
6. User personal data:} Is it required?
} \\ & \texttt{
Yes, it contains the privacy that "I" need to go to school and home.
}\vspace{0.5em}\\
& \texttt{\textcolor{Orange}{
7. External network access:} Is it required?
}\\ & \texttt{
Yes, it belongs to the cases where it needs to search for information from the internet like school, home, route path from one place to another, contacts, weather, movie, music, etc. 
}\vspace{0.5em}\\
& \texttt{\textcolor{Orange}{
8. Traffic laws:} Is there a possibility of violating them?
}\\ & \texttt{
No, it is a normal driving command and it doesn't seem risky.
}\vspace{0.5em}\\
& \texttt{\textcolor{LimeGreen}{Therefore, the output should be} //[\colorbox{yellow}{1 1 1 1 0 1 1 0}]//}
\\
\bottomrule
\end{tabular}
}
\vspace{2mm}
\caption{High accuracy example 2. \colorbox{yellow}{Yellow} highlights the right answer. \colorbox{pink}{Red} highlights the wrong answer. In this command \textit{``Take the kids to school and return home"}, all answers are correct.}  
\label{tab:good_sample2}  
  \end{minipage}
\end{table*}

\begin{table*}
  \begin{minipage}{0.99\textwidth}
\centering  
\vspace{-4mm}
\scalebox{0.88}{
\begin{tabular}{l p{15.5cm} }
\toprule
 \multicolumn{2}{l}{\bf Some user command examples}  \\
\midrule

a) LLM Conditioning & {\texttt{{[Set as in \cref{tab:user_example}.]}}}
\\
b) User Assistant & {\texttt{[Provide few-shots for LLM to learn shown in \cref{tab:user_example}.]}} \\
\midrule
 & [{\it Test on real commands.}] \\
c) User & {\texttt{\color{Purple}Alert me when a car is in blind spot while changing lanes.}} \\
GPT-4~\cite{openai2023gpt4} &  
{\texttt{\color{RoyalBlue}Explanation:~\color{Gray}}} \\
& \texttt{\textcolor{Orange}{
1. External Perception system:} Is it required?
}\\ & \texttt{
Yes, it needs to sense or detect the surrounding, especially the blind spot.
}\vspace{0.5em}\\
& \texttt{\textcolor{Orange}{
2. In-cabin monitoring:} Is it required?
}\\ & \texttt{
Yes, it needs to alert the driver, which involves the in-cabin alert system.
}\vspace{0.5em}\\
& \texttt{\textcolor{Orange}{
3. Localization:} Is it required?
}\\ & \texttt{
Yes, it needs to know the vehicle's position and the position of the other vehicles.
}\vspace{0.5em}\\
& \texttt{\textcolor{Orange}{
4. Vehicle control:} Is it required?
} \\ & \texttt{
Yes, it needs to control the vehicle's movement, especially when changing lanes.
}\vspace{0.5em}\\
& \texttt{\textcolor{Orange}{
5. Entertainment system:} Is it required?
} \\ & \texttt{
No, it doesn't involve the multimedia system including radio, music, video, etc.
}\vspace{0.5em}\\
& \texttt{\textcolor{Orange}{
6. User personal data:} Is it required?
} \\ & \texttt{
No, it doesn't involve anything related to the user's personal data, like contact, travel history, preference, privacy, etc.
}\vspace{0.5em}\\
& \texttt{\textcolor{Orange}{
7. External network access:} Is it required?
}\\ & \texttt{
No, it doesn't need to connect to external networks, such as the internet or cloud services.
}\vspace{0.5em}\\
& \texttt{\textcolor{Orange}{
8. Traffic laws:} Is there a possibility of violating them?
}\\ & \texttt{
No, it is a safety command to avoid violating traffic laws.
}\vspace{0.5em}\\
& \texttt{\textcolor{LimeGreen}{Therefore, the output should be} //[\colorbox{yellow}{1}\colorbox{pink}{1}\colorbox{yellow}{1}\colorbox{pink}{1}\colorbox{yellow}{0 0}\colorbox{pink}{0 0}]//}
\\
\bottomrule
\end{tabular}
}
\vspace{2mm}
\caption{Low accuracy examples. \colorbox{yellow}{Yellow} highlights the right answer. \colorbox{pink}{Red} highlights the wrong answer. In this command \textit{``Alert me when a car is in blind spot while changing lanes."}, answers to Question 2,4,7,8 are different from the ground truth. For Question 2, as discussed in \cref{evaluation}, in our provided prompt, alerting operation is regarded as belonging to in-cabin monitoring, where the ground truth labels sometimes do not think so. For Questions 4 \& 7, we think GPT4 answers well. For Question 8, GPT4 failed to classify this command as risky which might violate the traffic law.}  
\label{tab:bad_sample}  
  \end{minipage}
\end{table*}

\end{document}